\documentclass[11pt,a4paper]{article}
\usepackage[hyperref]{ranlp2023}
\usepackage{times}
\usepackage{latexsym}

\usepackage{microtype}

\aclfinalcopy %
\def\aclpaperid{29} %

\usepackage{graphicx}
\usepackage{multirow}

\title{Exploiting Primacy Effect To Improve Large Language Models}

\author{Bianca Raimondi \\
  University of Bologna, Italy \\
  \texttt{\normalsize bianca.raimondi3@unibo.it} \\\And
  Maurizio Gabbrielli \\
  University of Bologna, Italy \\
  \texttt{\normalsize maurizio.gabbrielli@unibo.it} \\}

\date{}

\begin{document}
\maketitle
\begin{abstract}
Large Language Models (LLMs) have become essential in many Natural Language Processing (NLP) tasks, 
leveraging extensive pre-training and fine-tuning to achieve high accuracy. 
However, like humans, LLMs exhibit biases, particularly positional biases such as primacy and recency effects, which
can influence the accuracy of the answers. 
The primacy effect—where items presented first are more likely to be remembered or selected—plays a key role in Multiple Choice Question Answering (MCQA), where the order of answer options can affect prediction outcomes.
This study focuses on primacy bias in fine-tuned LLMs: We first show that fine-tuning amplifies this bias, 
probably due to exposure to human-like patterns. Hence, we strategically leverage this effect by reordering response
options based on semantic similarity to the query, without requiring knowledge of the correct answer. 
Our experimental results show that this approach
significantly improves performance in MCQA.
More generally, our findings underscore the dual nature of biases as both challenges and opportunities, 
offering insights for bias-aware model design and NLP applications.
\end{abstract}

\section{Introduction}\label{sec:introduction}
\begin{figure}[t]
    \includegraphics[width=\linewidth]{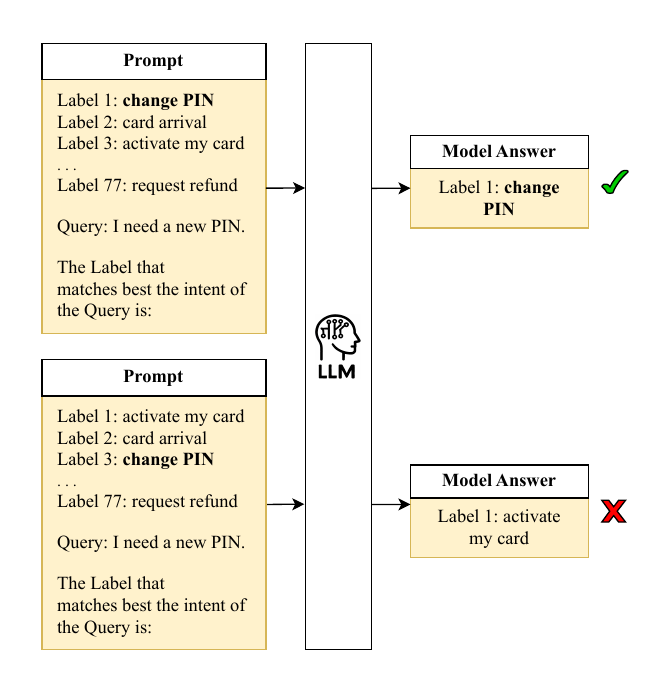}
    \caption{Given a Query (in this case, "I need a new PIN") and a set of 77 Options, the model must select the unique correct label (in this case, "Change PIN"). Due to the Primacy effect, the model tends to answer correctly only when the correct label is placed in the first positions.}
    \label{fig:sample}
\end{figure}
Large Language Models (LLMs) have emerged as powerful tools for a wide range of Natural Language Processing (NLP) tasks due to their remarkable accuracy, stemming from extensive pre-training on diverse corpora and fine-tuning on task-specific datasets.

However, despite their impressive capabilities, LLMs exhibit cognitive biases similar to those observed in humans, which can influence their decision-making processes~\cite{itzhak2024instructed,janik2023aspects,tjuatja2024llms}.
Particularly notable are positional biases such as Primacy and Recency effects, which tend to favour respectively the first and last items in a sentence~\cite{guo2024serialpositioneffectslarge,wang-etal-2023-primacy}.
These biases pose a challenge in tasks such as multiple-choice question answering (MCQA), where the placement of answer options can disproportionately influence the model's predictions, potentially leading to incorrect results~\cite{pezeshkpour-hruschka-2024-large,zheng2024large}.

This study focuses on the Primacy bias in LLMs within the context of MCQA. We first show that fine-tuned models have 
stronger positional biases than their pre-trained counterparts. We hypothesize that this is because fine-tuning amplifies this bias by exposing LLMs to repetitive human patterns.

Then, instead of considering these biases as a limitation, as usually done in the literature~\cite{wang-etal-2023-primacy}, we strategically leverage them to improve task performance. Specifically, we propose a technique that reorders the answer options based on their semantic similarity to the question, aligning them in descending order to exploit the primacy bias effectively. This approach is response independent; i.e., it enhances predictive accuracy without requiring prior knowledge of the correct answer.

Our approach is motivated by a simple yet effective intuition: if models are more likely to select early options, then arranging the most semantically relevant candidates near the top can guide the model toward more accurate predictions. In contrast to traditional approaches that attempt to neutralize bias through prompt engineering, our method embraces and exploits it. This perspective shifts the focus from bias mitigation to bias harnessing, opening new avenues for performance optimization in biased decision environments.

Our experimental results show that this approach significantly improves performance, particularly in fine-tuned models where the primacy bias is more pronounced. Since the models are also affected by the Recency effect, which is somehow the dual of the Primacy effect, we also consider an alignment of the answers options in ascending order. We show that also in this case, we have a gain in accuracy, and we then consider a combination of the two techniques. 

Finally, we show that the benefits of our technique are consistent across different model architectures and multiple MCQA datasets, demonstrating its robustness. These findings have broader implications for designing model-aware evaluation strategies and for creating more interpretable model behaviors in decision-making scenarios.

By addressing the dual nature of positional biases - as both challenges and opportunities - this work contributes to a deeper understanding of LLM behaviour and to the definition of strategies for bias-aware model design and application.

The structure of our work is organized as follows: Section~\ref{sec:related_work} provides a comparative overview of related work, Section~\ref{sec:methodology} details our experimental setup, Section~\ref{sec:results_and_discussion} presents and discusses our results, and Section~\ref{sec:conclusion_and_future_work} concludes with directions for future research.

For more details, the code of our work can be found here\footnote{https://github.com/biancaraimondi/PrimacyEffect.git}.

\section{Related Work}\label{sec:related_work}
Several recent studies have explored the phenomenon of positional biases in LLMs, revealing significant insights into their nature and implications.
Pezeshkpour et al.~\cite{pezeshkpour-hruschka-2024-large} show LLM performance fluctuates significantly when answer choices are reordered in MCQA, especially in cases of uncertainty between top answer options.

In the literature, two primary perspectives explain the origins of positional bias. The first attributes this effect to issues within the model's architecture. The second perspective suggests that the bias arises from the data used during fine-tuning, particularly human-labeled datasets. This latter viewpoint forms the basis for the first part of our work: testing positional bias in both pre-trained and fine-tuned versions of the models.

Research supporting the architectural hypothesis \cite{2023arXiv231001427P, guo2024serialpositioneffectslarge, DBLP:journals/corr/abs-2407-01100} emphasizes the role of components like the Positional Encoding mechanism in Transformers—such as RoPE~\cite{su2024roformer} used in the Llama model~\cite{touvron2023llama}—and the attention mechanism in shaping positional bias.

Studies attributing the problem to human-labeled data used during fine-tuning \cite{wang-etal-2023-primacy, janik2023aspects, 10.1162/tacl_a_00638, itzhak2024instructed} highlight their thesis testing models such as ChatGPT~\cite{chatgpt} after the Reinforcement Learning with Human Feedback (RLHF) phase.

An important contribution in this direction has been made by ~\citet{itzhak2024instructed}.
In this work, the cognitive bias is different from that in our study but achieves the same results as ours, suggesting that fine-tuning either introduces cognitive-like biases or reinforces them if they are already present.

Some research was conducted on how to reduce the positional bias in LLMs.
\citet{eicher2024compensatory} approach involves introducing intermediary steps, or guard rails, in prompts to moderate this bias, but with mixed success.
\citet{zheng2024large} show that LLMs have a bias toward selecting certain option IDs (like "Option A") in MCQA due to token bias.
To address this, the authors propose a debiasing method that adjusts predictions by estimating the model's prior bias.
\begin{figure*}[t]
    \includegraphics[width=\linewidth]{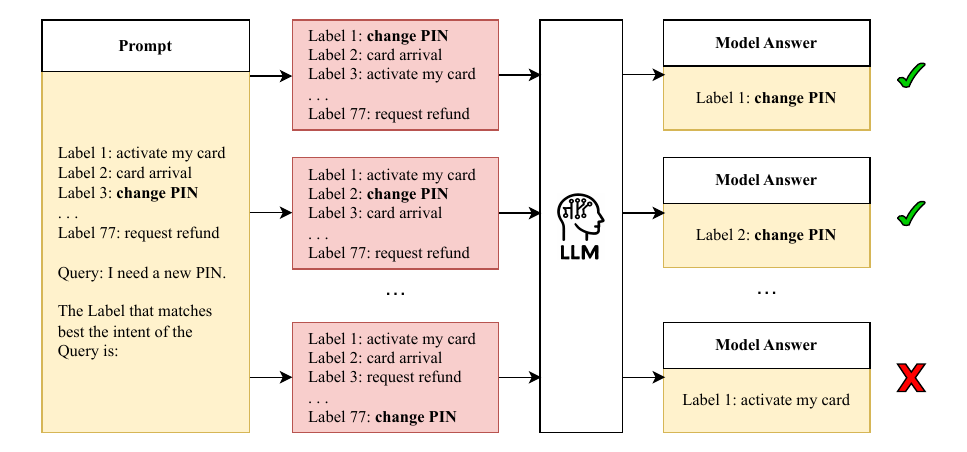}
    \caption{An example of the evaluation process for Primacy bias detection. Each query is fixed, and the target label is systematically shuffled across all possible positions. For each position, the model's prediction is recorded, and its correctness is determined by comparing it to the target label. Most of the time, when the target label is placed in the first positions, the model predicts the correct answer, confirming the presence of the Primacy bias.}
    \label{fig:pretrained_finetuned}
\end{figure*}
Finally, \citet{tjuatja2024llms} show that changes in query structure do not alter model responses, unlike human counterparts.

Collectively, these studies offer diverse perspectives on positional biases in LLMs, shedding light on their origins, their impact on task performance, and strategies for mitigation through architectural adjustments or training data interventions. In contrast to previous efforts, our work introduces a novel technique aimed at leveraging this bias to enhance model accuracy, rather than attempting to eliminate it.

\section{Methodology}\label{sec:methodology}
In the following methodology section, we describe our experimental framework for evaluating positional biases in LLMs in the MCQA task.
This includes details about the datasets and models used, the comparison between pre-trained and fine-tuned versions, and the development of a training-free reordering technique to exploit the Primacy bias.
\subsection{Datasets and Models}
In our experiments, we tested several LLMs on the MCQA task, where the model selects the correct label from a given query and multiple options. We used three available datasets: CLINC~\cite{larson-etal-2019-evaluation}, BANKING~\cite{casanueva-etal-2020-efficient}, and HWU~\cite{liu2021benchmarking}.
Each dataset contains a different number of samples, where a query is presented with multiple options, only one of which is correct. The options remain the same across samples, while the query changes. This setup allows us to analyze model behavior when the options are shuffled. An example can be found in Figure~\ref{fig:sample}.
For CLINC, we used a reduced version with 3,750 samples and 150 possible options. BANKING consists of 3,080 samples with 77 options, and HWU has 3,080 samples with 54 options.
These datasets vary not only in the number of options but also in the complexity and domain-specific nature of the queries, enabling us to examine model performance across different contexts.
We tested Mistral 7B versions~\cite{jiang2023mistral} and various models from the Llama family: Llama3 8B~\cite{dubey2024llama}, Llama2 13B, and 7B~\cite{touvron2023llama}.
\begin{figure*}[t]
    \centering
    \includegraphics[width=.99\textwidth]{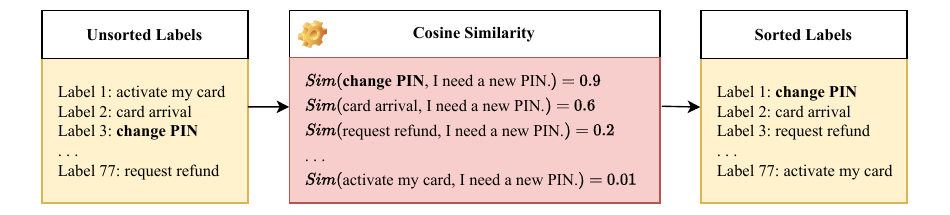}
    \caption{An example of the ranking process of options based on their similarity to the query. The cosine similarity is computed between query and option embeddings. The options are then ranked in descending order of similarity, with the most similar option placed first.}
    \label{fig:CosineSimilarity}
\end{figure*}
Models of varying scales and designs were used to assess positional bias and the effectiveness of reordering.

\subsection{Pre-trained vs Fine-tuned versions}\label{sec:pretrained_finetuned}
To conduct the tests, we fixed each query and systematically shuffled the target label across all possible positions, from 0 to the total number of options, as shown in Figure~\ref{fig:pretrained_finetuned}.
For each position, we recorded the model's prediction and noted whether it matched the target label or not.
This procedure allowed us to isolate the influence of label position from other confounding variables such as content or semantics.

The models evaluated included both base pre-trained checkpoints and their instruction-tuned or RLHF-enhanced variants.
All models were run in zero-shot settings with fixed prompts to avoid introducing variability from dynamic prompt construction.
To mitigate any effects from randomness, we used a temperature of 0 and disabled sampling.

A key methodological distinction between the pre-trained and fine-tuned models lies in their exposure to human-aligned instruction data.
While pre-trained models operate purely based on next-token prediction over generic text corpora, fine-tuned models have been further optimized on curated datasets that contain instructional prompts and human-preferred responses.
This additional training phase is hypothesized to introduce positional heuristics that reflect human annotation tendencies, such as favoring early options.
To support a fair comparison, we ensured that both types of models received the same input queries and answer shuffling schemes.
In doing so, we established a controlled framework for attributing behavioral differences specifically to the fine-tuning stage.
In Section~\ref{sec:results_and_discussion}, we then plot the results comparing pre-trained and fine-tuned versions of the same model to show differences in the presence of positional bias.
These comparisons highlight the extent to which task-specific supervision during fine-tuning can inadvertently reinforce cognitive-like biases in LLM behavior.

\subsection{Reordering}\label{sec:reordering}
The main goal of our work is to develop a technique that reorders options to place the correct label in the first positions, leveraging the Primacy effect. To achieve this, we designed a shuffling method that accomplishes this without requiring prior knowledge of the target label. This is particularly useful in cases where datasets are unlabeled or when the true output is unknown.
In contrast to supervised approaches that depend on annotated data or reinforcement learning, our method offers a lightweight and generalizable alternative.
Our approach is inspired by the idea that semantically similar options are more likely to be correct, and that LLMs, due to the Primacy bias, are more likely to select earlier options regardless of semantic content. By merging these two tendencies, we maximize the chance that the semantically best option is also the one most likely to be selected.

We sorted the options based on their similarity to the query.
For each sample, we computed the mean cosine similarity between the embeddings of the query and the options.
Since cosine similarity is computed per token rather than for entire queries or options, we followed the intuition of~\citet{DBLP:journals/corr/abs-2407-01100}, aggregating similarity scores across tokens.
This approach implicitly aligns semantic proximity with positional advantage, which is particularly effective given the positional bias we aim to exploit.

While this approach does not guarantee that the correct label will always be top-ranked, our empirical evaluation shows that the correlation between semantic similarity and correct labeling is strong enough to offer substantial gains across different datasets and model configurations.
Moreover, because the technique requires only forward passes through a frozen encoder to obtain embeddings, it is computationally efficient and can be scaled to large datasets with minimal resource overhead.

\begin{figure}[t]
    \includegraphics[width=\linewidth]{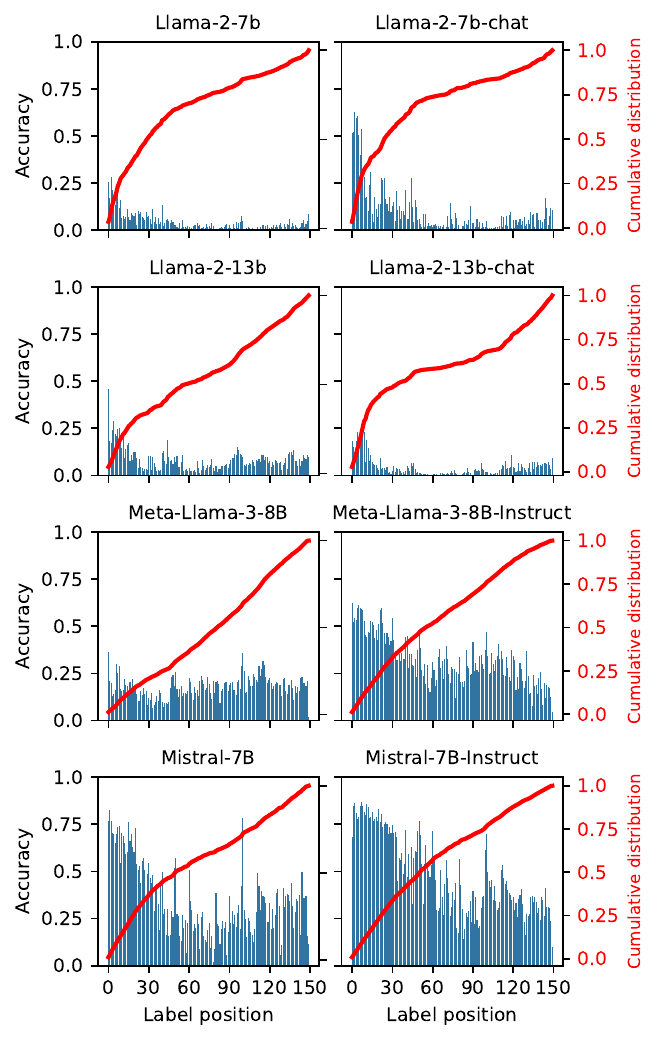}
    \caption{Comparison of Primacy bias between pre-trained and fine-tuned versions of models for the CLINC dataset. The x-axis represents the position of the target label, and the y-axis shows accuracy over all the samples. The Cumulative distribution, represented by the red line in the plots, shows the proportion of total accuracy accumulated as the label position increases. Fine-tuned models demonstrate a stronger Primacy bias, with higher accuracy for labels in early positions.}
    \label{fig:cumulative_distribution}
\end{figure}
To detail our process mathematically:
\begin{enumerate}
    \item Token-wise Cosine Similarity: For each token \( t_o \) in the option embedding \( o \) and each token \( t_q \) in the query embedding \( q \), we compute the cosine similarity: \[\ cos\_sim(t_o, t_q) = \frac{t_o \cdot t_q}{\|t_o\| \|t_q\|}\] where \( t_o \cdot t_q\) denotes the dot product of token embeddings \( t_o \) and \( t_q \), and \( \|t_o\| \) and \( \|t_q\| \) are their magnitudes.
   \item Aggregation of Token Similarities: We then compute the mean of these similarities over all token pairs, defining the overall similarity between the query and each option:
   \[
   Sim(O, Q) = \frac{\sum_{t_o \in O} \sum_{t_q \in Q} cos\_sim(t_o, t_q)}{|O| \cdot |Q|}
   \]
   where \( O \) is the set of tokens in the option and \( Q \) is the set of tokens in the query.
\end{enumerate}
\begin{figure}[t]
    \includegraphics[width=\linewidth]{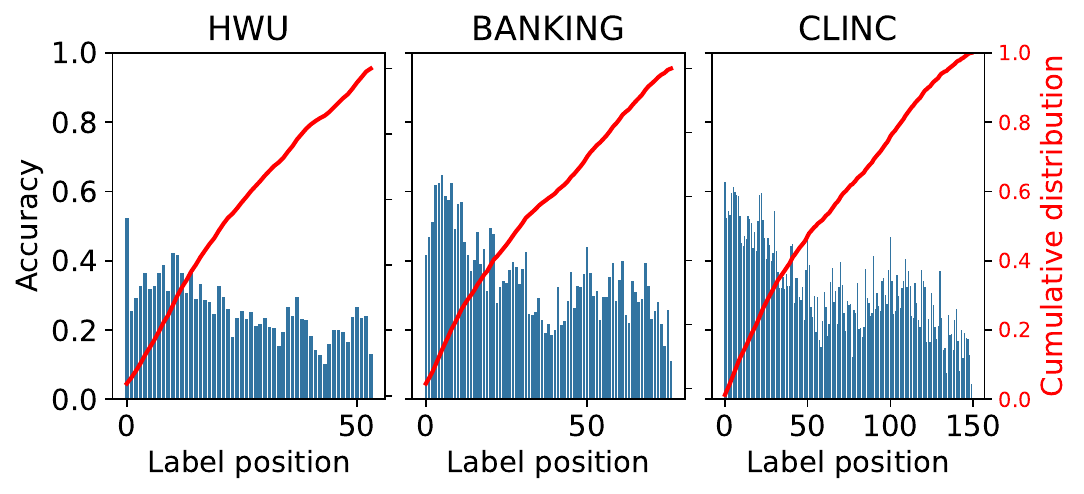}
    \caption{Comparison of Primacy bias in Llama3-8B-Instruct across three datasets with varying numbers of labels. The bias is less pronounced in the HWU dataset (54 labels) and more pronounced in the CLINC dataset (150 labels), demonstrating that the Primacy effect intensifies as the number of labels in the prompt increases.}
    \label{fig:cumulative_distribution_labels_number}
\end{figure}
The result, which provides a single mean similarity score per option, ranks the options in descending order, placing the most similar option first.
This similarity-driven reordering can be viewed as a soft heuristic proxy for relevance, aligning semantically coherent choices with the model’s attention priority.
In some cases, we observed that even when the correct label was not ranked first, it was consistently moved toward the top half of the options list, significantly improving its chances of selection compared to a random or original ordering.

Empirically, we show that this ordering significantly increases the likelihood of correct predictions by systematically leveraging LLMs’ tendency to favor early options.
Our overall work is shown in Figure~\ref{fig:CosineSimilarity}.

\section{Results and Discussion}\label{sec:results_and_discussion}
To evaluate positional bias, we compared the performance of pre-trained and fine-tuned versions (Instruction Tuning and RLHF) of four models: Llama2-7B, Llama2-13B, Llama3-8B, and Mistral-7B. As explained in Section~\ref{sec:methodology}, we systematically shuffled the target label across all positions and recorded the accuracy at each. The results show a consistent trend: fine-tuned versions exhibit a more pronounced Primacy bias than their pre-trained counterparts.
Llama3-8B exhibited the strongest Primacy bias among the tested models, with the fine-tuned version showing a pronounced preference for target labels at the beginning of the option list.
Figure~\ref{fig:cumulative_distribution} compares the pre-trained and fine-tuned versions of each model.

Here, the x-axis represents the position \(p\) of the target label, while the y-axis shows accuracy, i.e., the number of samples where the model's prediction matches the target label at \(p\).
The accuracy is computed using the formula: 
\begin{equation}  
\label{eq:accuracy}  
A_{M,S}(p) = \frac{|\{s \ | \ l_s = M(s_{p}), \: \forall s \in S\}|}{|S|} 
\end{equation}  
where \(M\) is the model, \(S\) is the set of samples, \(l_s\) is the target label for sample \(s\), and \(M(s_{p})\) is the model's label prediction for sample \(s\) with the target label positioned at \(p\).  
Results for each model are provided in Table~\ref{tab:models_accuracy}.

Interestingly, despite exhibiting a stronger Primacy bias, the fine-tuned Llama3-8B consistently outperforms its pre-trained counterpart across all label positions. This observation, as seen in Fig.\ref{fig:cumulative_distribution} and Table\ref{tab:models_accuracy}, indicates that Primacy bias does not necessarily harm model performance. Instead, it appears that Instruction Tuning improves the model's general ability to map inputs to correct outputs, even if positional biases are amplified as a side effect. This suggests that bias and accuracy are not strictly inversely related in this context.

To validate our result, we tested the models on the three datasets mentioned in Section~\ref{sec:methodology}.
The tests reveal that the Primacy effect is more pronounced when the prompt contains more labels. Figure~\ref{fig:cumulative_distribution_labels_number} compares the results of Llama3-8B-Instruct across the datasets, showing lower bias in the HWU dataset, which has 54 labels, and higher bias in the CLINC dataset, which has 150 labels; thus showing a more pronounced bias in favor of larger prompts.

Given the Primacy bias, our work focused on exploiting it to improve model accuracy.
As introduced in Section~\ref{sec:methodology}, we achieved this by sorting the options in descending order of similarity to the query.
\begin{table}
\centering
\begin{tabular}{cl|ccc}
\hline %
\multicolumn{1}{c}{}
& \multicolumn{1}{c|}{\textbf{Model}}
& \textbf{NoSort}
& \textbf{Sort}
& \textbf{Oracle}
\\
\hline
\multirow{4}{*}{PT} &
Llama-2-7B & 0.03 & 0.07 & 0.26\\
& Llama-2-13B & 0.08 & 0.16 & 0.46\\
& Llama-3-8B & 0.20 & 0.27 & 0.36\\
& Mistral-7B & 0.41 & 0.53 & 0.77\\
\hline
\multirow{4}{*}{FT} &
Llama-2-7B & 0.12 & 0.19 & 0.51\\
& Llama-2-13B & 0.02 & 0.13 & 0.18\\
& Llama-3-8B & 0.37 & 0.49 & 0.63\\
& Mistral-7B & 0.39 & 0.50 & 0.68\\
\hline
\end{tabular}
\caption{Accuracy of pretrained (PT) and finetuned (FT) models under our descent sorting strategy (\textit{Sort}) for CLINC dataset.
Our technique (\textit{Sort}) increases models accuracy relative to the \textit{NoSort} baseline.
The \textit{Oracle} serves to represent an idealised approach that always places the answer in the first position.}
\label{tab:models_accuracy}
\end{table}
Table~\ref{tab:models_accuracy} presents the results of our models comparisons under three conditions: without reordering (\textit{NoSort}), using our proposed technique (\textit{Sort}), and leveraging an idealized upper bound represented by an \textit{Oracle}.
The \textit{Oracle} simulates a system that always ranks the correct answer first, providing an estimate of the maximum achievable accuracy.
By comparing our technique to the \textit{Oracle}, we can evaluate the effectiveness in leveraging bias and quantify the performance improvement relative to this theoretical upper limit.

The results show a clear improvement when models employ descending reordering, emphasizing the benefits in increasing accuracy.
Variability in performance across models could be influenced by differences in training data or architecture, indicating that reordering benefits may vary depending on the model or application.
These findings highlight the potential of descending reordering to enhance performance, particularly in ranking or classification tasks where precision is critical.

\begin{figure}[t]
    \includegraphics[width=\linewidth]{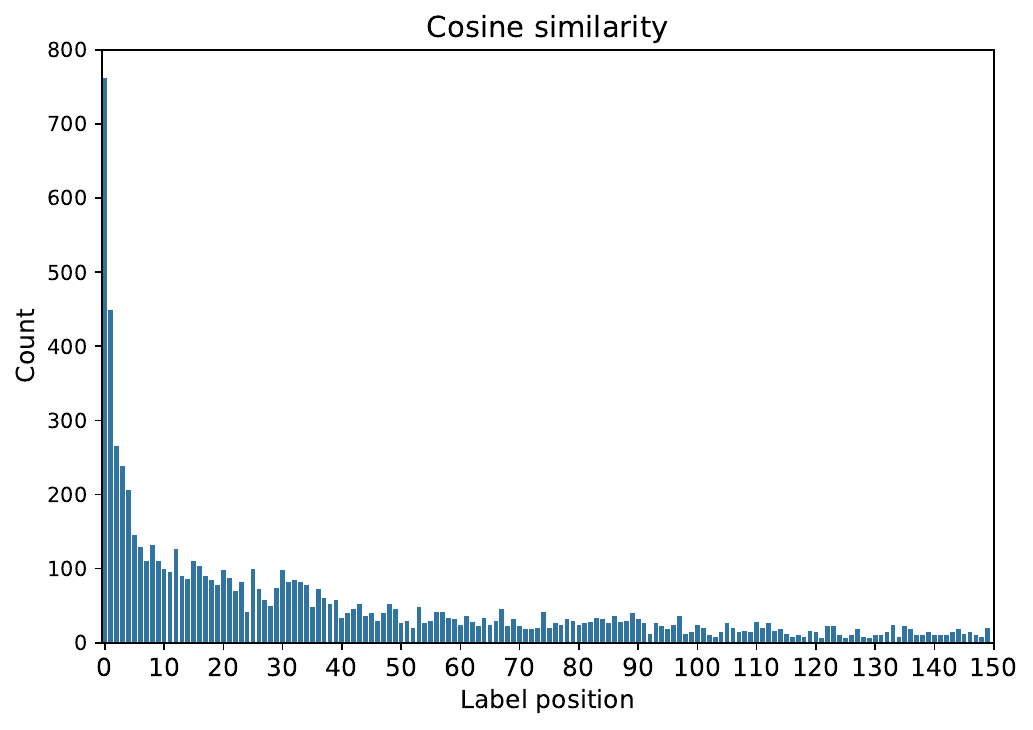}
    \caption{Distribution of the target label's position after sorting options based on their similarity to the query in Llama3-8B-Instruct model. The barplot shows the number of samples in CLINC dataset where the correct label is placed at position \(x\). The results indicate that the target label is often placed in the top positions (i.e., where \(x\) is small), demonstrating that the proposed method can be used effectively to leverage Primacy bias to improve model accuracy.}
    \label{fig:distribution_target_label}
\end{figure}
We also assess whether the target option has been placed in the first positions.
As shown in Figure~\ref{fig:distribution_target_label}, testing the Llama3-8B-Instruct model with the CLINC dataset demonstrates that our technique consistently places the target label in the top positions.
The barplot shows a higher value for labels placed in the first positions, indicating that our method is effective.

To further analyze the effectiveness of our similarity-based reordering strategy in leveraging positional bias, we evaluated the performance of Llama-3-8B and Mistral-7B across all three datasets.

\begin{table}[!ht]
\centering
\begin{tabular}{llccc}
\hline
\multicolumn{1}{l}{\textbf{FT Model}} & \multicolumn{1}{l}{\textbf{Dataset}} &
\textbf{Top-1} & \textbf{Top-10} \\
\hline
Llama-3-8B & \multirow{2}{*}{CLINC} & 10.16 & 33.97 \\
Mistral-7B &  & 17.72 & 42.12 \\
\hline
Llama-3-8B & \multirow{2}{*}{BANKING} & 9.24 & 46.54 \\
Mistral-7B &  & 21.40 & 62.47 \\
\hline
Llama-3-8B & \multirow{2}{*}{HWU} & 12.00 & 30.27 \\
Mistral-7B &  & 29.94 & 47.15 \\
\hline
\end{tabular}
\caption{Percentage of samples where the correct target label appears within the top-1 and top-10 positions after reordering.}
\label{tab:topk_positions}
\end{table}
Table~\ref{tab:topk_positions} reports the percentage of samples in which the correct label appears within the top-k positions after reordering options by descending cosine similarity to the query.
Results show that both models benefit significantly from this reordering. Mistral-7B consistently achieves higher coverage than Llama-3-8B across all datasets and top-k thresholds, suggesting that it is better aligned to leverage similarity-based ordering for narrowing down candidate labels.

In CLINC, for instance, Mistral-7B places the correct label within the top-1 position in 17.72\% of samples, compared to 10.16\% for Llama-3-8B. The advantage extends to broader thresholds: at top-10, Mistral-7B achieves 42.12\% coverage versus 33.97\% for Llama-3-8B. A similar trend is observed on BANKING and HWU, with Mistral-7B demonstrating a particularly strong top-1 coverage of 29.94\% on HWU, outperforming Llama-3-8B by nearly 18 percentage points.

These coverage results should not be interpreted as direct accuracy comparisons between models, but rather as an indication of how effectively each model, when paired with our reordering method, surfaces the correct label within the top portion of the list. The high coverage at top-10 suggests that the similarity metric effectively ranks relevant candidates early, providing a practical mechanism to guide model attention in tasks involving long candidate lists.

In summary, these findings validate the effectiveness of our similarity-based reordering in improving the positioning of correct answers within candidate lists, especially when combined with models like Mistral-7B that are better calibrated for this type of heuristic.

\subsection{Ablation studies}
\paragraph{Similarity Metrics}
\begin{table}[!ht]
\centering
\begin{tabular}{llc}
\hline
\multicolumn{1}{l}{\textbf{FT Model}} & \multicolumn{1}{l}{\textbf{Metric}} &
\textbf{Acc.} \\
\hline
\multirow{4}{*}{Llama-3-8B} & Manhattan Distance & 0.374 \\
 & Euclidean Distance & 0.405 \\
 & Cosine Similarity & 0.490 \\
 & Sentence Transformer & \textbf{0.634} \\
\hline
\multirow{4}{*}{Mistral-7B} & Manhattan Distance & 0.403 \\
 & Euclidean Distance & 0.405 \\
 & Cosine Similarity & \textbf{0.502} \\
 & Sentence Transformer & 0.439 \\
\hline
\end{tabular}
\caption{Impact of different similarity metrics on accuracy for CLINC150.}
\label{tab:similarity_metrics}
\end{table}
To compute similarity between queries and options, we adopted the cosine similarity metric, as it is a standard approach for comparing textual embeddings.

To assess this empirically, we conducted an ablation study comparing cosine similarity from the model embedder against alternative distance measures (Euclidean and Manhattan Distances) and the \texttt{all-MiniLM-L6-v2} Sentence Transformer\footnote{\href {https://huggingface.co/sentence-transformers}{https://huggingface.co/sentence-transformers}.} embeddings.

The results in Table~\ref{tab:similarity_metrics} show that sentence transformers achieve significantly higher accuracy with Llama-3-8B than with cosine similarity (0.634 versus 0.490), supporting the idea that contextual advantage could potentially be a substitute. However, for Mistral-7B, cosine similarity outperforms sentence transformers (0.502 vs. 0.439), suggesting that optimal similarity methods may be model-dependent, possibly due to differences in their internal representation spaces.

Regarding computational efficiency, we chose token-average similarity initially due to its integration simplicity within the LLM’s own embedding space, avoiding external models and inference overhead. Sentence transformers, while providing stronger semantic representations, introduce additional computational costs due to the need for separate model forward passes for each candidate option.

\begin{figure}[!ht]
    \includegraphics[width=\linewidth]{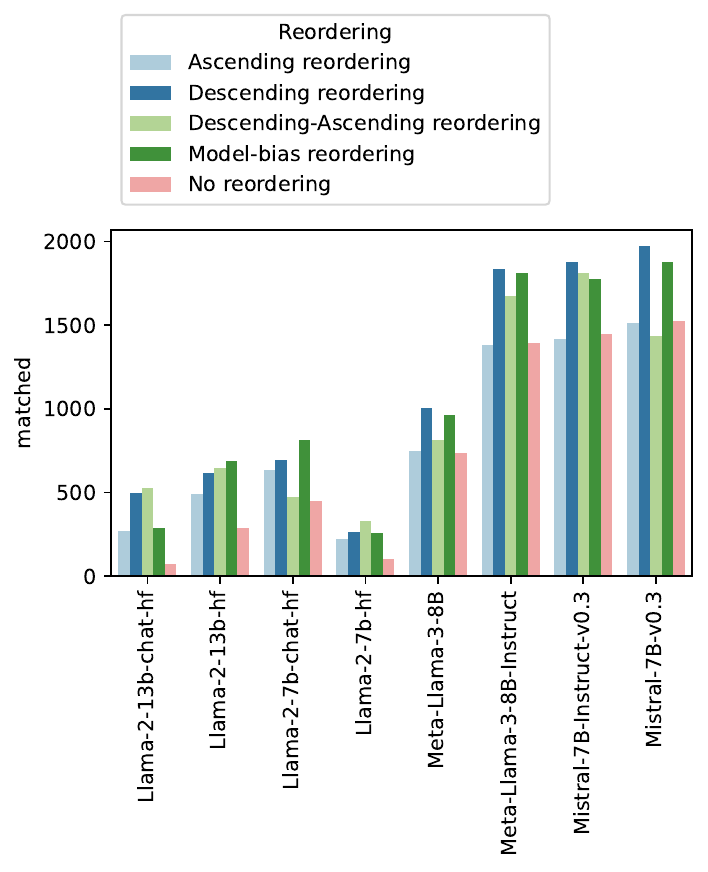}
    \caption{Barplots showing, for each model, the number of matched samples answers with options sorted in \textit{Descending} and \textit{Ascending} order based on their similarity to the query. The \textit{Descending-Ascending} order exploits both Primacy and Recency effects, while the \textit{Model-bias} order leverages the model-specific bias. The results confirm that descending order, which places the target label in the first positions, leads to higher accuracy compared to other sorting mechanisms, at least for models with reasonable accuracy. This supports the hypothesis that exploiting the Primacy bias improves model performance.}
    \label{fig:reordering_overview_onecolumn}
\end{figure}
\paragraph{Ordering}Since our first idea was to improve the model by placing the target label in the first positions to exploit the Primacy effect, we sorted the options in descending order of similarity to the query.

Given that LLMs are also influenced by the Recency Effect, where they tend to answer correctly when the correct label appears last, sorting the options in ascending order might improve accuracy.
For models like Llama-2, our results support this hypothesis, confirming the presence of Recency bias and the effectiveness of using an ascending order.

We also implemented two other techniques aimed at combining the two biases.
The first one duplicates the options and creates a concatenated list where the options appear first in descending order and then in ascending order.
The idea behind this process is to try to exploit both Primacy and Recency effects at the same time.
The second technique exploits the bias associated with the specific model: it stores, for each position, the accuracy of the models using results from our first work explained in section~\ref{sec:pretrained_finetuned}.
It then places the options in the stored positions, starting by placing the option with the highest option-query similarity in the position with the highest accuracy value.
Other options are then placed using the same logic until the last free position is reached.

Figure~\ref{fig:reordering_overview_onecolumn} compares the results of all techniques. For models like Llama-2, the last two techniques show improvement.
However, for models that perform well before reordering (shown in the \textit{No reordering} columns), the descending order yields the best results.
This highlights the importance of Primacy bias when interacting with the model.

\section{Conclusion and Future Work}\label{sec:conclusion_and_future_work}
We investigated positional bias in MCQA, showing that fine-tuned LLMs exhibit Primacy effect. Leveraging this, we introduced a simple, training-free method that reorders answer options by their similarity to the query, significantly improving accuracy without requiring labeled data. Our approach works well across different models and datasets, proving robust and generalizable results.

Rather than treating positional bias as a flaw, we show its potential as a performance lever. This opens new paths for bias-aware inference strategies that align cognitive patterns with task goals.

Future work will refine this method, extend it to other biases (e.g., emotional \cite{mozikov2024eai}), and explore adaptive reordering and integration with prompt engineering.

\section*{Limitations}\label{sec:limitations}
Our findings may be subject to certain limitations. The Primacy effect may vary across LLM architectures and fine-tuning methods, potentially reducing the impact of our approach. Our target label shuffling may not generalize to tasks like multi-hop reasoning, where position plays a lesser role. Reliance on cosine similarity assumes embedding relevance aligns with semantic correctness, which may not hold in cases requiring a deeper understand. 

Moreover, our method's robustness across domains—especially in low-resource languages, multimodal inputs, or ambiguous labels—remains to be validated. Finally, our evaluation on fixed-format MCQA datasets may not capture real-world query complexity.

\section*{Acknowledgments}
We acknowledge CINECA\footnote{\href {https://www.cineca.it}{https://www.cineca.it}.} for the availability of computing resources and support.

This work was supported by Future Al Research (FAIR) PE01, SPOKE 8 on PERVASIVE AI funded by the National Recovery and
Resilience Plan (NRRP).

\bibliographystyle{acl_natbib}
\bibliography{ranlp2023}

\end{document}